# Deep Learning Approach in Automatic Iceberg – Ship Detection with SAR Remote Sensing Data

*Cheng Zhan; Licheng Zhang; Zhenzhen Zhong; Sher Didi-Ooi, University of Bristol; Youzuo Lin, Los Alamos National Laboratory; Yunxi Zhang, UT Health; Shujiao Huang; Changchun Wang*

## Summary

Deep Learning is gaining traction with geophysics community to understand subsurface structures, such as fault detection or salt body in seismic data. This study describes using deep learning method for iceberg or ship recognition with synthetic aperture radar (SAR) data. Drifting icebergs pose a potential threat to activities offshore around the Arctic, including for both ship navigation and oil rigs. Advancement of satellite imagery using weather-independent cross-polarized radar has enabled us to monitor and delineate icebergs and ships, however a human component is needed to classify the images. Here we present Transfer Learning, a convolutional neural network (CNN) designed to work with a limited training data and features, while demonstrating its effectiveness in this problem. Key aspect of the approach is data augmentation and stacking of multiple outputs, resulted in a significant boost in accuracy (logarithmic score of 0.1463). This algorithm has been tested through participation at the Statoil/C-Core Kaggle competition.

## Introduction

Drifting icebergs present a significant threat to ships operating and oil rig activities including in the so-called Iceberg Alley, offshore of East Coast of Canada – where the legendary Titanic ship met its fate.

Advancement of earth observation with satellites using remote sensing has opened a new avenue of earth science research through offering tremendous amount of possibilities for better understanding of the earth's environment and assisting in sound decision making. Synthetic Aperture Radar (SAR) provides a continuous monitoring with high-resolution two-dimensional satellite imagery which are independent from daylight, cloud coverage, fog, rain and various weather conditions (Moreira et al., 2013). The C-band system is a dual-polarization radar which transmits a horizontal electromagnetic signal and is able to receive the backscattered responses in both horizontal (HH) and vertical polarization (HV). The amplitude and phase of backscattered responses is dependent on the physical (i.e. surface roughness, geometry) and the electrical (i.e. dielectric) properties of the imaged target. Today, more than 15 spaceborne SAR satellites are operating for numerous applications including sea ice zone and land surface monitoring (both dynamic and static).

The training data provided include the 75 x 75 pixel images with associated features: HH, HV, and target output of whether the image is a ship or an iceberg. Some images are easier to classify, while others are not straightforward (see Figure 3). The goal of this study is to extract or generate valuable features to further enhance the effectiveness of the algorithm to automatically answer the question, is it an iceberg or a ship?

In the following sections, we will first formalize the iceberg and ship delineation problem given by Statoil and C-Core for a Kaggle competition. Next, we describe the proposed algorithm and results, and provide the discussion and conclusion.

## Data Analysis and Augmentation

Global properties of the images were statistically explored – minimum, maximum, mean, median, first quartile, third quartile, standard deviation of the two bands HH and HV - and consequently became new features for the training data. The correlations between them are presented in Figure 1. Newly generated features were implemented with decision tree based model (gradient boosting method) which achieved a decent result, with the *logloss* value of 0.21.

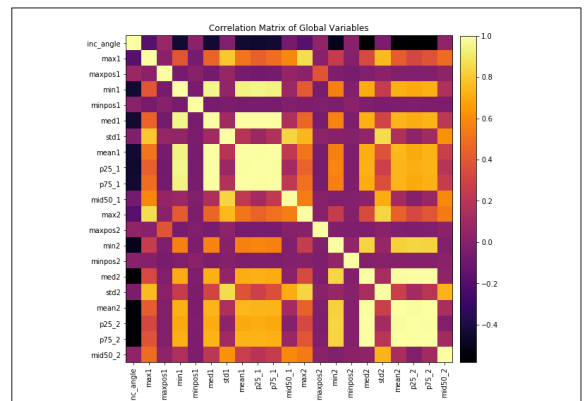

Figure 1. Correlation matrix of HH and HV band features including min, max, mean, median, first quartile, third quartile, and standard deviation.

# Deep learning in Automatic Icebergs Detection Using Remote Sensing Satellite Data

A deep learning training model presented a challenge in its build due to the limited set of data. While more data is desirable, this is limited by computing power and resource. The augmentation here is accomplished by a variety of image transformations, including rotation, reflection, smoothing, first and second derivatives, gradient and Laplacian. The first pass experiment began by gradually increasing the size of the data while fixing the underlying model, and the testing error and training error decreased (Figure 2).

Further of increase in data size is seen to drive the convergence of the training loss and test loss curves – indicating the model being less overfitting, enabling us to generalize broader test data.

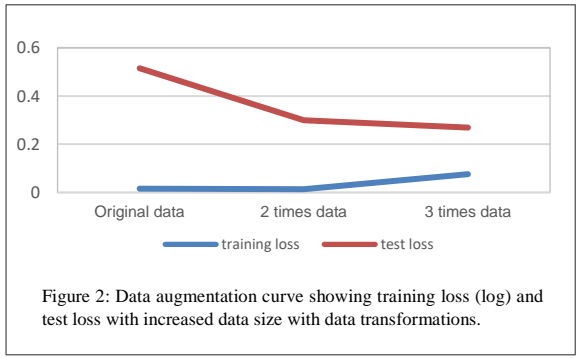

Figure 2: Data augmentation curve showing training loss (log) and test loss with increased data size with data transformations.

An example of the data augmentation code using Keras:

```
from keras.preprocessing.image import ImageDataGenerator

traindatagenerator=
ImageDataGenerator(width_shift_range=0.1,
height_shift_range=0.1, rotation_range=15)
```

### Feature Engineering

Figure 3 highlights the different SAR images that were provided with visually straightforward differences that may help discern between icebergs and ships, and thus providing additional features and further improve our classification model. The first is the color composite images which shows iceberg having more yellow than ship. Other observation is size, whereby icebergs tend to be larger than ships, allowing us to include additional features, difference (HH – HV) and ratio (HH/HV).

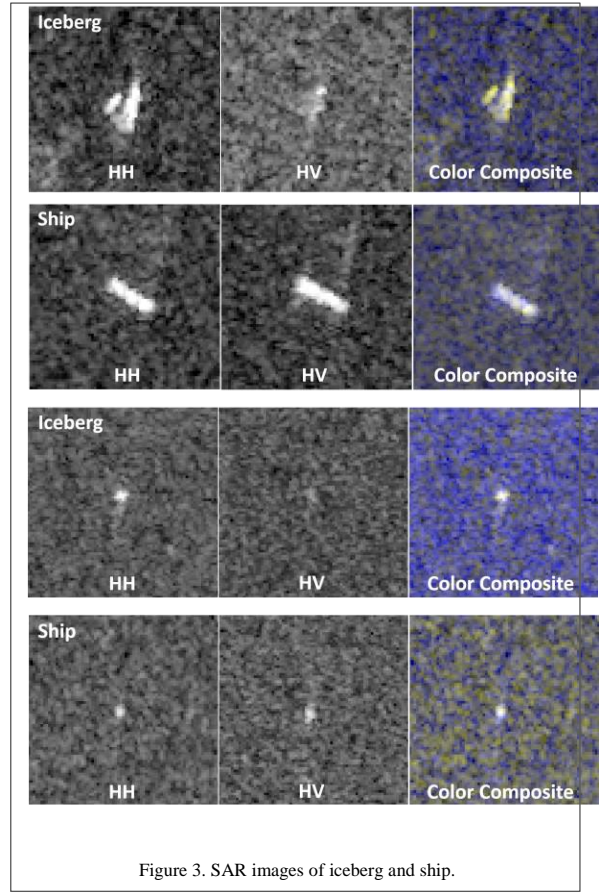

Figure 3. SAR images of iceberg and ship.

A crucial strategy for feature engineering is leveraging subject matter expertise of the domain, for example the incidence angle (SAR range 20° - 45°), radar beam what is perpendicular to the surface, and backscatter intensity (reflectivity), which is decreases with increase incidence angle. The smoothness of the object affects the backscatter intensity (Figure 4). Backscatter coefficient, is the conventional measure of the strength of radar reflected by distributed scatter, and is normalized by the cosine of the incidence angle.

# Deep learning in Automatic Icebergs Detection Using Remote Sensing Satellite Data

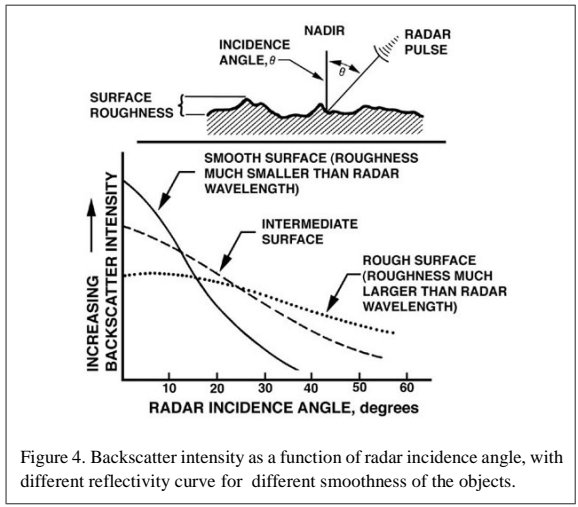

Figure 4. Backscatter intensity as a function of radar incidence angle, with different reflectivity curve for different smoothness of the objects.

**Deep Learning/Transfer Learning**

The deep learning framework includes splitting the training data into training and validation sets to 4:1 using the function *train_test_split* from python library *sklearn*. The trained model performance was evaluated on the validation set using the *accuracy* metric, with the loss function *logloss*, which quantifies the accuracy by penalizing false classifications. A useful strategy was found with representing the learning rate as a variable of the performance. For example, if the monitored performance on the validation set does not improve after certain number of epochs, the learning rate is reduced by 90%. This is done to avoid the solution to skip the optimum and jump back and forth when it approaches the optimum. Hence reducing the learning rate helps the model to get closer to the target. On the other hand, if the learning rate is reduced too fast and too quickly, the optimization can becomes extremely inefficient.

In the context of machine learning, hyperparameters are a set of parameters whose values are determined prior to the commencement of the learning process. In short, the values of other parameters are derived through training. Optimizing hyperparameters for a learning algorithm is important. For deep learning framework, calibration of hyperparameters include optimization algorithm, learning rate, drop off rate, number of hidden layers, number of units in each layer etc.

The best deep learning models are at present still a 'black box' in which there are no scientific explanations on why certain neural network architectures are superior than others. The next best approach to this 'black box' in finding a good model is to evaluating against the test set. The number of layers and units in each hidden layer is challenging to tune, which necessitates the adoption of existing and proven model design as the starting point, with further domain-based modifications. Thus, we will be focusing on testing of different optimization algorithms and different methods of setting the learning rate. The architect used for hyperparameterisation is displayed using Keras visualization (Figure 5).

Comparison of different optimization methods and tests of hyperparameters have led us to use Adam optimization algorithm with initial learning rate 0.001. Some attractive benefits of using Adam include: ease of implementation, computationally efficient, appropriate for problems with very high noise or sparse gradients. The training-validation loss, training-validation accuracy, amd confusion matrix of the prediction on the validation set are shown in Figure 6. This achieved accuracy 89%.

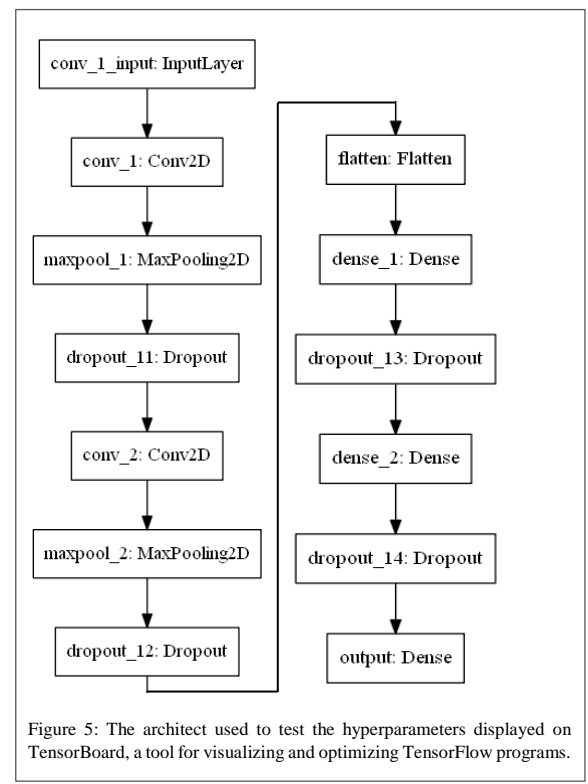

Figure 5: The architect used to test the hyperparameters displayed on TensorBoard, a tool for visualizing and optimizing TensorFlow programs.

# Deep learning in Automatic Icebergs Detection Using Remote Sensing Satellite Data

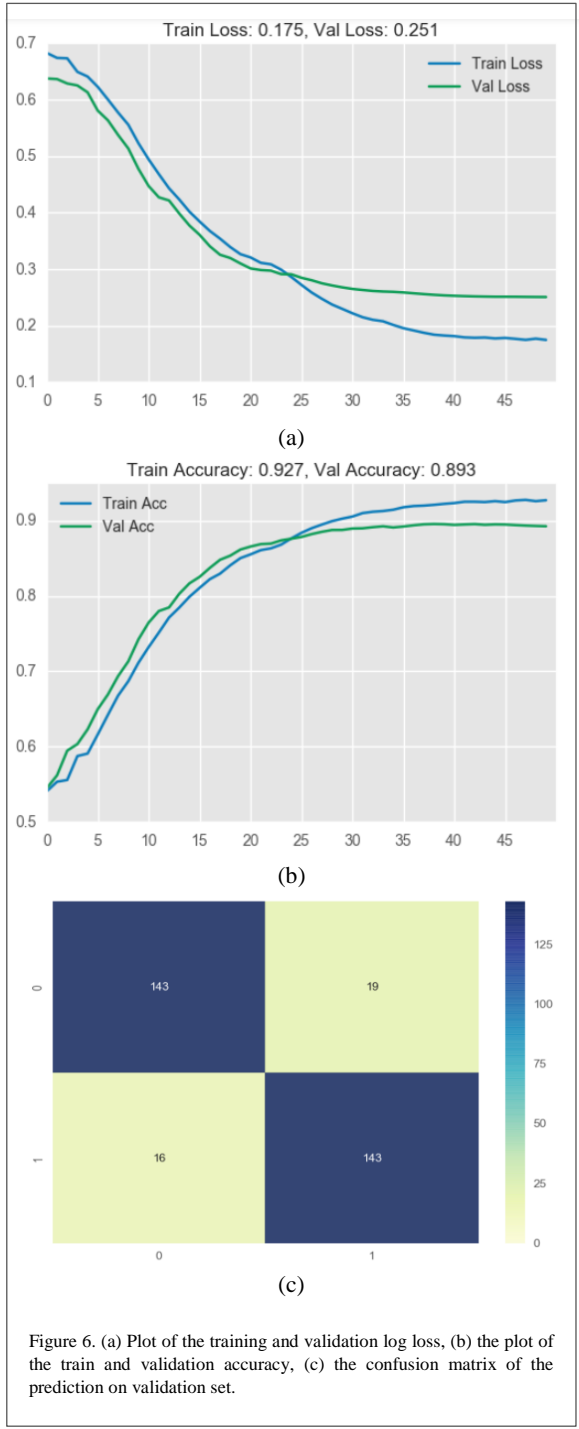

Figure 6. (a) Plot of the training and validation log loss, (b) the plot of the train and validation accuracy, (c) the confusion matrix of the prediction on validation set.

The main challenge of this classification problem is with limited data available, even after data augmentation using different transformations. One solution is to leverage transfer learning method, whose effectiveness has been demonstrated (Huang, et al., 2017). The key component to this is to 'borrow' knowledge learned from sufficient unlabeled SAR transferable to labelled SAR target data. In the workflow, there are two pathways: classification and reconstruction with a feedback bypass. Many images were used to train reconstruction pathway with stacked convolutional auto-encoders, and the pre-trained layers were then reused via 'transfer knowledge' to the classification task.

**Discussion and Conclusion**

We have been able to apply the deep learning algorithm to automaticly recognize whether an image is a Ship or an Iceberg. Inspired by the NMO processing from seismic data, one method to utilize the incident angle is by normalization. Using transformation, the images are standardized to same incident angle, say 0°. This may help the process of discrimination between ship and iceberg.

After creating several uncorrelated models, the other method is to further boost the results is by stacking or assembling. There are many methods to proceed with such technique and the principle is to use the wisdom of the crowd.

So far, our best score of *logloss* is 0.1463, and the associated accuracy is higher than 0.95. The ranking so far is top 11%, and we should be able to push the score further.